%% file: ijcai21.tex
\documentclass{article}
\usepackage{ijcai21}
\usepackage{tikz}
\usepackage{todonotes}

\usepackage{times}
\usepackage{soul}
\usepackage{url}
\usepackage[hidelinks]{hyperref}
\usepackage[utf8]{inputenc}
\usepackage[small]{caption}
\usepackage{graphicx}
\usepackage{amsmath}
\usepackage{amsthm}
\usepackage{booktabs}
\usepackage{algorithm}
\usepackage{algorithmic}
\usepackage{xspace}
\usepackage{cleveref} 
\newcommand*\frontaleye{%
       \scalebox{0.21}{
\tikzset{every picture/.style={line width=0.75pt}}   
\begin{tikzpicture}[x=0.75pt,y=0.75pt,yscale=-1,xscale=1]
\draw  [fill={rgb, 255:red, 0; green, 0; blue, 0 }  ,fill opacity=1 ] (84.83,179.7) .. controls (84.92,169.21) and (100.66,160.85) .. (119.99,161.01) .. controls (139.32,161.18) and (154.92,169.81) .. (154.83,180.3) .. controls (152.72,170.91) and (137.95,163.54) .. (119.97,163.39) .. controls (101.99,163.23) and (87.1,170.35) .. (84.83,179.7) -- cycle ;
\draw  [fill={rgb, 255:red, 0; green, 0; blue, 0 }  ,fill opacity=1 ] (102.39,180.93) .. controls (102.39,171.24) and (110.25,163.39) .. (119.93,163.39) .. controls (129.62,163.39) and (137.47,171.24) .. (137.47,180.93) .. controls (137.47,190.61) and (129.62,198.47) .. (119.93,198.47) .. controls (110.25,198.47) and (102.39,190.61) .. (102.39,180.93) -- cycle ;
\draw  [draw opacity=0][fill={rgb, 255:red, 255; green, 255; blue, 255 }  ,fill opacity=1 ] (107,174) .. controls (107,171.24) and (109.24,169) .. (112,169) .. controls (114.76,169) and (117,171.24) .. (117,174) .. controls (117,176.76) and (114.76,179) .. (112,179) .. controls (109.24,179) and (107,176.76) .. (107,174) -- cycle ;
\draw  [fill={rgb, 255:red, 0; green, 0; blue, 0 }  ,fill opacity=1 ] (154.65,179.7) .. controls (154.65,190.58) and (139.02,199.41) .. (119.74,199.41) .. controls (100.46,199.41) and (84.83,190.58) .. (84.83,179.7) .. controls (85.71,190.15) and (101,198.47) .. (119.74,198.47) .. controls (138.48,198.47) and (153.77,190.15) .. (154.65,179.7) -- cycle ;
\end{tikzpicture}}
\,}

\definecolor{positive}{HTML}{1155cc}
\definecolor{negative}{HTML}{cc0000}

\input{math_commands.tex} 

\newcommand{\pting}{pretraining\xspace}
\newcommand{\Pting}{Pretraining\xspace}

\newcommand{\pted}{pretrained\xspace}

\crefname{section}{\S}{\S\S}
\Crefname{section}{\S}{\S\S}
\crefname{table}{Tab.}{}
\crefname{figure}{Fig.}{}
\crefname{algorithm}{Algorithm}{}
\crefname{equation}{eq.}{}
\crefname{appendix}{App.}{}
\crefname{prop}{Proposition}{}
\crefformat{section}{\S#2#1#3} 

\title{A Primer on Contrastive \Pting in Language Processing:\\ Methods, Lessons Learned 
 and Perspectives}
\author{
Nils Rethmeier$^{1,2}$
\and
Isabelle Augenstein$^2$
\affiliations
$^1$German Research Center for AI, Berlin, Germany\\
$^2$University of Copenhagen, Copenhagen, Denmark\\
\emails
nils.rethmeier@dfki.de, augenstein@di.ku.dk
}

\begin{document}
\maketitle
\begin{abstract}
Modern natural language processing (NLP) methods employ self-supervised \pting objectives such as masked language modeling to boost the performance of various application tasks. These \pting  methods are frequently extended with recurrence, adversarial or linguistic property masking, and more recently with contrastive learning objectives.
Contrastive self-supervised training objectives enabled recent successes in image representation \pting by learning to contrast input-input pairs of augmented images as either similar or dissimilar.
However, in NLP, automated creation of text input augmentations is still very challenging because a single token can invert the meaning of a sentence.
For this reason, some contrastive NLP \pting methods contrast over input-label pairs, rather than over input-input pairs, using methods from Metric Learning and Energy Based Models.
In this survey, we summarize recent self-supervised and supervised contrastive NLP \pting methods and describe where they are used to improve language modeling, few or zero-shot learning, \pting data-efficiency and specific NLP end-tasks. We introduce key contrastive learning concepts with lessons learned from prior research and structure works by applications and cross-field relations. Finally, we point to open challenges and future directions for contrastive NLP to encourage bringing contrastive NLP \pting closer to recent successes in image representation \pting.
\end{abstract}

\section{Introduction}
\begin{figure}[t]
    \centering
    \includegraphics[width=.98\linewidth, trim=0 0 0 0, clip]{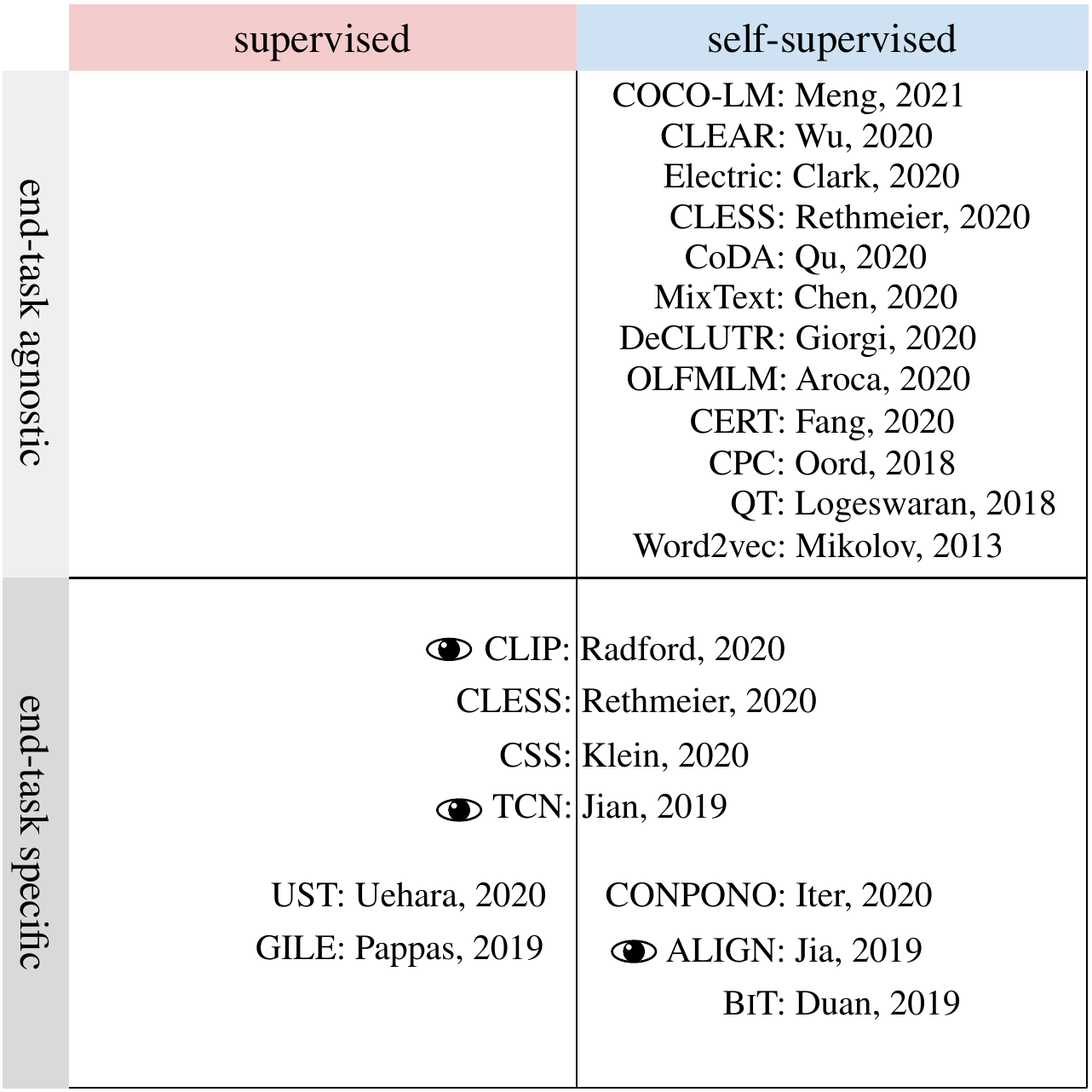}
    \caption{\textbf{Types of contrastive \pting} and works that fall within these categories. \protect\frontaleye marks text-image contrastive works.}\label{fig:type_of_contrastive_models}
\end{figure}
Current downstream machine learning applications heavily rely on the effective \pting of representation learning models. Contrastive learning is one such technique which enables \pting of general or task-specific data encoder models in a supervised or self-supervised fashion to increase the downstream performance of language or image representations. While contrastive \pting in computer vision has enabled the recent successes in self-supervised image representation \pting, the benefits and best practices of contrastive \pting in natural language processing (NLP) are still comparatively less established \cite{CTLSURVEY}. 
However, there is a first line of works on contrastive NLP methods which show strong performance and data-efficiency benefits of (self-)supervised contrastive NLP \pting as illustrated in \cref{fig:type_of_contrastive_models}. For example, supervised contrastive \pting enables zero-shot prediction of unseen text classes and improves few-shot performance \cite{GILE}.
Moreover, task-agnostic self-supervised contrastive \pting systems have been shown to improve language modeling \cite{QuickThoughts,EBM_LM_PRETRAINING,CLEAR,DECLUTR}, while \cite{rethmeier2020longtail} develop a data-efficient contrastive \pting method for improved zero-shot and long-tail learning. Others propose task-specific contrastive self-supervision for pronoun disambiguation \cite{NABIKLEIN}, discourse representation learning \cite{CONPONO}, text summarization \cite{CT_text_summarization} and other NLP tasks, as we will describe in \cref{sec:clt_sl_ssl}.

\paragraph{Contributions:}
In this primer to contrastive \pting, we therefore summarize recent (self-)supervised contrastive NLP \pting methods and describe how they enable zero-shot learning and improve language modeling, few-shot learning, \pting data-efficiency or rare event prediction. We cover basic concepts and crucial design lessons of contrastive NLP, while detailing the resulting benefits such as zero-shot prediction and efficient training. 
Then, we structure existing research as supervised or self-supervised contrastive \pting and explain connections to energy based models (EMBs), since many works refer to EBMs. Finally, we point out open challenges and outline future and underrepresented research directions in contrastive NLP \pting.

\section{Contrastive Learning Concepts and Benefits}
At the core of contrastive methods is the idea of learning to contrast between pairs of similar and dissimilar data points. A pair of similar data points is called a positive sample if both data points are different representations or views of the same data instance. Negative samples are pairs where the two data points are of different data instances. For contrastive learning, such data points can either be input-input $(x_i,x_j)$ or  input-label $(x_i,y_j)$ pairs. While contrastive computer vision methods learn from input-input (image-image) pairs $(x_i,x_j)$ \cite{CTLSURVEY,BIGSIMCLR}, NLP methods  additionally use input-output (text, label) pairs $(x_i,y_c)$. Here $x_i$ are input text embeddings, while $y_c$ are label embeddings of a short text that describes a label, i.e.\ an extreme summarization of the input text to get two views of said text. 

\subsection{Noise Contrastive Estimation (NCE)}\label{sec:NCE}
Noise contrastive estimation is the objective used by most contrastive learning approaches within NLP. Thus, we briefly outline its main variants and the core ideas behind them, while pointing to \cite{NCE}\footnote{\url{https://vimeo.com/306156327} talk by \cite{NCE}.} for detailed, yet readily understandable explanations of the two main NCE variants. Both variants can intuitively be understood as a sub-sampled softmax with \textcolor{negative}{$K$} negative samples \textcolor{negative}{$a_i^-$} and one positive sample \textcolor{positive}{$a_i^+$}. The first variant expresses NCE as a binary objective (loss) in the form of maximum log likelihood, where only \textcolor{negative}{$K$} negatives are considered.
\begin{align}
\begin{aligned}
L_B(\theta,\gamma) & = log\, \sigma(s(x_i,\textcolor{positive}{a_{i,0}^+};\theta), \gamma)  \label{eqn:NCE_AS_MLE} \\
 & \hphantom{=} + \sum_{k=1}^{\textcolor{negative}{K}} log(1-\sigma(s(x_i,\textcolor{negative}{a_{i,k}^-};\theta), \gamma) 
\end{aligned}
\end{align}
Here, $s(x_i,{a_{i,\circ}};\theta)$ is a scoring or similarity function that measures the compatibility between a single text input $x_i$ and another sample $a_{i,\circ}$. As mentioned above, the sample can be another input text or an output label (text), thus modeling NLP tasks as `text-to-text' prediction similar to language models. The similarity function is typically a cosine similarity, a dot product or a logit (unscaled activation) produced by a input-sample matcher sub-network \cite{rethmeier2020longtail}. The $\sigma(z,\gamma)$ is a scaling function, which for use in \cref{eqn:NCE_AS_MLE} is typically the sigmoid $\sigma(z)=\exp(z-\gamma)/(1+\exp(z-\gamma))$ with a hyperparameter $\gamma\geq0$ (temperature), that is tuned or omitted depending on the way that negative samples \textcolor{negative}{$a_i^-$} are attained.

The other NCE objective learns to rank a single positive pair $(x_i,\textcolor{positive}{a_{i,0}^+})$ over \textcolor{negative}{$K$} negative pairs $(x_i,\textcolor{negative}{a_{i,k}^-})$:
\begin{align}
\begin{aligned}
L_R(\theta) & = log \frac{e^{\textstyle{{\bar{s}(x_i,\textcolor{positive}{a_{i,0}^+};\theta)}}}}
{e^{\textstyle{{\bar{s}(x_i,\textcolor{positive}{a_{i,0}^+};\theta)}}} + \sum_{k=1}^{\textcolor{negative}{K}}e^{\textstyle{{\bar{s}(x_i,\textcolor{negative}{a_{i,k}^-};\theta)}}}} 
\end{aligned}
\end{align}
Here, to improve $L_R$ or $L_B$ performance, \cite{NCE} propose a regularized scoring function $\bar{s}(x_i,a_{i,\circ}) = s(x_i,a_{i,\circ}) - log\, p_\mathcal{N}(a_{i,\circ})$ that subtracts the probability of the current sample $a_{i,\circ}$ under a chosen noise distribution $p_\mathcal{N}(a_{i,\circ})$. In practice, the noise distribution can be set to 0 \cite{NCE_unnormalized,CLEAR,rethmeier2020longtail} to save on computation. To robustly learn word embeddings, $p_\mathcal{N}(a_{i,\circ})$ can be set as the word probability $p_{word}$ in a corpus \cite{W2V}, or as the probability of a sequence under a language model $p_{LM}$ \cite{CONTRASTIVE_TEXT_GENERATION}, when learning contrastive sequence prediction.

\paragraph{Generalization to an arbitrary number of positives:} As \cite{supervisedContraLearn} mention, original contrastive formulations use only one positive pair per text instance (see e.g. \cite{W2V,QuickThoughts}), while more recent methods mine multiple positives or use multiple gold class annotation representations for contrastive learning \cite{rethmeier2020longtail,ContrastiveDataAugmentationPretraining}. This means that e.g. the positive term in \cref{eqn:NCE_AS_MLE} becomes $\sum^{P}_{p=1} log\, \sigma(s(x_i,\textcolor{positive}{a_{i,p}^+};\theta, \gamma))$ to consider \textcolor{positive}{$P$} positives.

\paragraph{Importance of negative sampling semantics and lessons learned:}
How positive and negative samples are generated or sampled is a key component of effective contrastive learning. \cite{ContrastiveLearningLimitations} prove and empirically validate that ``sampling more negatives improves performance, but only if they are sampled from the same context or block of information such as the same paragraph''. Such hard to contrast (classify) negatives, are sampled in most works \cite{W2V,ContrastiveLearningLimitations,rethmeier2020longtail,CONPONO}. Otherwise, performance can deteriorate due to weak contrast learning of conceptually related classes. Additionally, \cite{rethmeier2020longtail} find that both positive and negative contrastive samples from a long-tail distribution are essential in predicting rare classes and in substantially boosting zero-shot performance, especially over minority classes. \cite{W2V} undersample negatives of frequent words to stabilize \pting of word embeddings to a similar effect. Additional practical advice for negative sampling is mentioned in \ref{sec:SSL_CL}.

\subsection{Contrastive Learning as Mutual Information Maximization, Inverse Data Generation and Energy Based Models:}\label{sec:EBM}
Contrastive learning methods are closely related to at least four machine learning concepts. First, InfoNCE has been shown to maximize the lower bound of mutual information between different views of the data \cite{Oord,Contrastive_as_MInfo_max}. Second, \cite{CLT_INVERTS_DATA_GENERATION}, show that contrastive learning robustly inverts a data generation process ``by uncovering the true generative factors of variation underlying the observational data, even in practical cases, where most theoretical assumptions of the generation process are severely violated.''. Third, learning similarities in data connects contrastive learning to metric learning \cite{MetricLearningProblems}. Finally, many works describe contrastive learning as an Energy Based Model, EBM, and since this may initially be unfamiliar, we outline popular EBM variations for supervised and self-supervised contrastive text \pting below.

\begin{figure}[t]
    \centering
    \includegraphics[width=.77\linewidth, trim=0 0 0 0, clip]{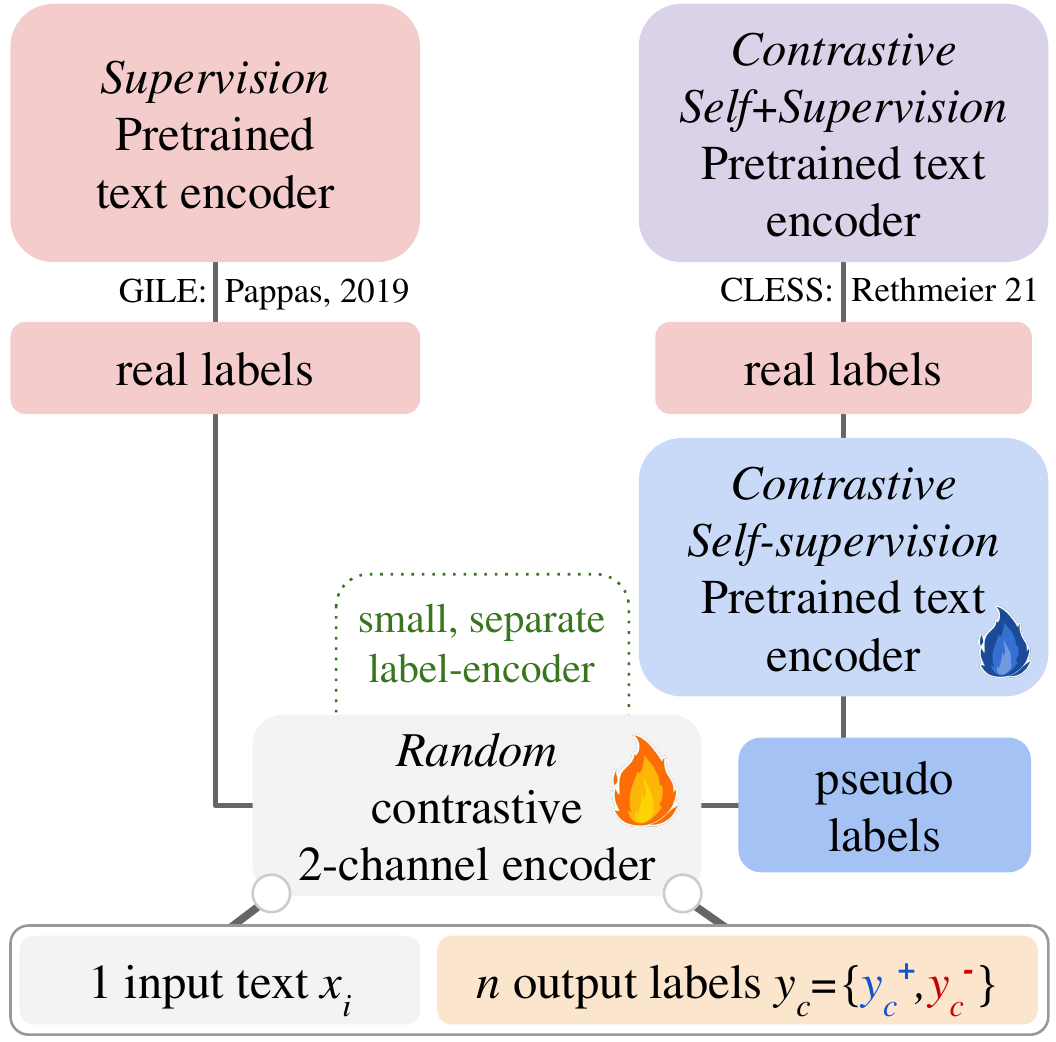} 
    \caption{\textbf{Contrastive input-output $(X,Y)$ \pting.} Texts and labels are encoded independently via a medium sized text encoder and a very small label-encoder. This encodes $1$ text for $n$ labels with minimal computation to enable large-scale $K$ negative sampling.}\label{fig:input-output}
\end{figure}

\paragraph{Input-output contrastive EBM:} The binary NCE variant from \cref{eqn:NCE_AS_MLE} is a special case of a ``Contrastive Free Energy'' loss as described in \cite{LECUN_EBM_AND_CONTRASTIVE} Fig. 6b or in \cite{EBM_AS_DISCRIMINATIVE_LOSS} Fig.\ 2 and Sec.\ 3.3 as the negative log-likelihood loss with negative sampling. \cite{LECUN_EBM_AND_CONTRASTIVE} originally state that an EBM $E$ learns the compatibility between input-output pairs $(x_i,y_c)$ with $x_i\in X$ and $y_c\in Y$ 
\begin{align}
    E(X,Y)\; \text{or}\; E(W, X, Y)  \label{eqn:IO}
\end{align}
where $W$, or $\theta$ in \cref{eqn:NCE_AS_MLE}, are model parameters that encode inputs $X$ and labels $Y$. Here, $X$ and $Y$ are views or augmentations of either the same data point (positives), or different data points (negatives). The energy function $E$ measures the compatibility between its parameters $(X,Y)$, where $E(\circ){=}0$ indicates optimal compatibility -- e.g.\ $E(X{=}Tiger, Y{=}felidae){=}0$ means $X$ and $Y$ match. Note that in the probabilistic framework $P(Y{=}felidae|X{=}Tiger,W){=}1$. Works which use input-output noise contrastive estimation are \cite{GILE,rethmeier2020longtail}, visualized in \cref{fig:input-output}. They encode an input text $x_i$ using a text-encoder $T$ and a label description $y_c$ via a separate label-encoder $L$ to then concatenate both into a single text input-output encoding pair $(T(x_i),L(y_c))$. Once encoded, the input-label pair similarity is learned via a binary NCE objective $L_B$ as in \cref{eqn:NCE_AS_MLE}. Compared to input-input models described below, these approaches allow for encoding a large number of augmented views, i.e.\ labels, very compute efficiently via a small label-encoder. This allows them to scale to large sample sizes of positives and negatives, which is crucial to successful contrastive learning. While \cite{GILE} use this formulation for supervised-only \pting on label encodings, \cite{rethmeier2020longtail} additionally sample input words $x_i\,{\in}\,X$ as pseudo-label encodings $y_c'{=}L(x_i)$ for efficient contrastive self-supervised \pting. Thus, the later approach unifies supervision and self-supervision as a single task of contrasting real-label encodings $L(y_c)$ or pseudo-label encodings $y_c'{=}L(x_i)$. The advantage of such methods is that once the NCE classifier is pretrained, it can be reused, i.e.\ zero-shot transferred, to any downstream task, without having to initialize a new classifier. In fact, unified prediction and zero-shot transfer are properties one would expect to have from \pting, since most NLP tasks fit into a `text-to-text' prediction description.
As a result of contrastive pseudo-labels, input-output methods enable efficient contrastive self-supervised \pting \cite{rethmeier2020longtail}, even on very small data, with commodity hardware, and without complicated mechanisms like cyclic learning rate schedules, residual layers, warmup, specialized optimizers or normalization which current large-data \pting approaches require as research summarized in \cite{BERT_PRETRAINI_DIFFICULTIES} shows. Finally, many input-input contrastive methods rely on re-\pting already otherwise \pted Transformer architectures \cite{CERT,CONTRASTIVE_TEXT_GENERATION,DECLUTR}, since encoding augmented inputs is costly in current Transformer architectures. 

\begin{figure}[t]
    \centering
    \includegraphics[width=.98\linewidth, trim=0 0 0 0, clip]{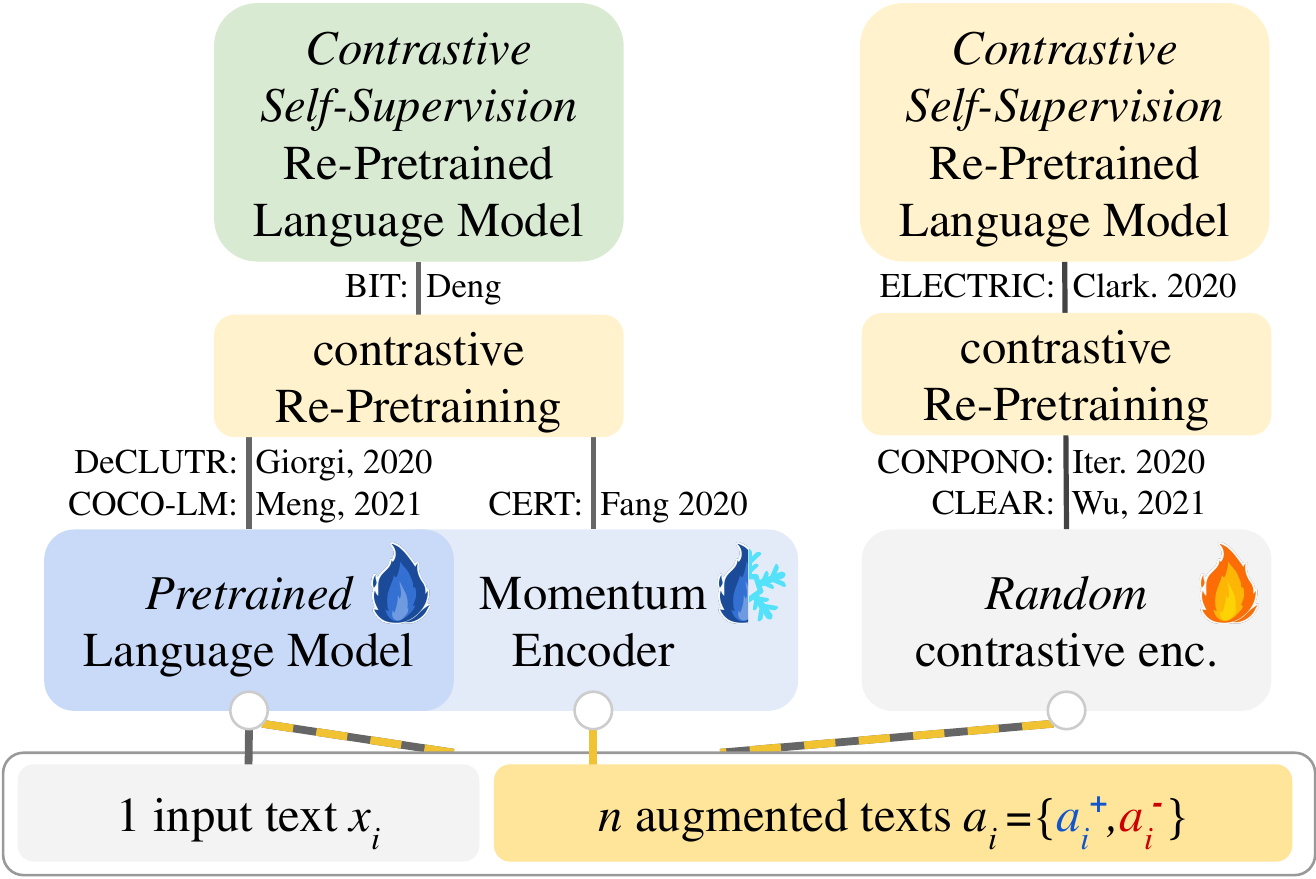} 
    \caption{\textbf{Contrastive input-input $(X,X')$ \Pting:} Input-input methods contrast an original text with augmented positive \textcolor{positive}{$a_i^+$} and negative \textcolor{negative}{$a_i^-$} texts $a_i\in X'$, which requires more computation than input-output methods. Achromatism compatible.}\label{fig:input-input}
\end{figure}

\paragraph{Input-input contrastive EBM:} 
Input-input methods contrast input texts $X$ from augmented input texts $X'$ rather than from labels $Y$ -- see \cref{fig:input-input}. For example, \cite{EBM_LM_PRETRAINING} replace a subset of input text words $x_{i,w}$ with other words $x_{i,w'}$ sampled from the vocabulary for self-supervised contrastive \pting. The original text $x_i$ is augmented into a text $a_i$ to provide a positive sample augment \textcolor{positive}{$a_i^+$} or a negative sample augment \textcolor{negative}{$a_i^-$}. Self-supervised \pting then contrasts pairs $(x_i,a_i)$ of original texts against augmented ones via the binary NCE as in \cref{eqn:NCE_AS_MLE}. Similar to the EBM in \cref{eqn:IO} this can be summarized as   
\begin{align}
    E(X,X')\; \text{or}\; E(W,X,X') 
\end{align}
As mentioned, current input-input contrast models are hampered by compute-intense augmentation encoding $W(a_i)$.

\paragraph{Contrastive \pting enables zero-shot learning, improves few-shot learning and increases parameter learning efficiency:}\label{sec:IO_II}
\cite{CLIP} replace a Transformer by a CNN to speed up self-supervised zero-shot prediction learning by a factor of 3, and add text contrastive \pting to speed up learning by another factor of 4. 
\cite{GILE} show that supervised contrastive \pting enables supervised zero-shot and improved few-shot learning. \cite{rethmeier2020longtail} run self-supervised contrastive \pting for unsupervised zero-shot prediction, i.e.\ without human annotations, and show that this boosts learning performance on long-tail classes. This is done while \pting on only portions of an already very small text collection of 6 to 60MB of \pting text. They also demonstrate that rather than adding more data during \pting, one can also increase self-supervised learning signals instead.

\section{Self- or Supervised Contrastive Pretraining}\label{sec:clt_sl_ssl}
The goal of contrastive \pting is to initialize model weights for efficient zero-shot transfer or fine-tuning to downstream tasks. \pting is either supervised or self-supervised. Supervised contrastive \pting methods use corpora of hand-annotated data such as paraphrased parallel sentences, textual labels or text summarizations to define text data augmentations for contrastive \pting. Self-supervised contrastive methods aim to scale \pting by contrasting automatically augmented input texts $X'$ or textual output pseudo-labels $Y'{\sim}P(X)$ -- see \cref{sec:EBM} for input-input vs.\ input-output contrastive methods.    
Both self-supervised and supervised contrastive methods are used to train language encoder models from scratch, or can `re-pretrain' or fine-tune an already otherwise \pted model such as a RoBERTa \cite{ROBERTA}. Below, we structure self- and supervised contrastive \pting by technique and application. 

\subsection{Self-supervised Contrastive Pretraining} \label{sec:SSL_CL}

\paragraph{Input-input contrastive text representation \pting via automated text augmentation:}
\cref{fig:input-input} compares methods that use input-input contrastive (EBM) learning as overviewed in \cref{sec:IO_II}. \cite{ContrastiveDataAugmentationPretraining} use a contrastive momentum encoder over combinations of recently proposed text data augmentations like ``cutoff, back translation, adversarial augmentation and mixup''. They find that mixing augmentations is most useful when the augmentations provide sufficiently different views of the data. Further, since constructing text augmentations which do not alter the meaning (semantics) of a sentence is very difficult, they introduce two losses to ensure both sufficient difference and semantic consistency of sentence augmentations. They define a consistency loss to guarantee that augmentations lead to similar predictions $y_c$ and a contrastive loss that makes augmented text representations $a_i$ similar to the original text $x_i$. To ensure that a sufficiently large amount of negative text augmentations are sampled, they use an augmentation-embedding memory bank. \cite{CERT} only use back-translation, \cite{CLEAR,COCOLM} investigate other sentence augmentation methods, \cite{DECLUTR} contrast text spans, \cite{EBM_LM_PRETRAINING,COCOLM} replace input words by re-sampling a language model and \cite{SentenceStructurePT} investigate contrastive sentence structure \pting. Finally, \cite{COCOLM} also contrasts cropped sentences after augmentation via word re-sampling.  

\paragraph{Contrasting Next or Surrounding Sentence (or Word) Prediction (NSP, SSP)}
Sentence prediction is a popular input-input contrastive method as in \cref{sec:IO_II}.
Next sentence prediction, NSP, and surrounding sentence prediction, SSP, take inspiration from the skip-gram model \cite{W2V}, where surrounding and non-surrounding words are contrastively predicted given a central word to learn word embeddings using an NCE \cref{sec:NCE} variant \cite{W2V}. Methods mostly differ in how they sample positive and negative sentences, where negative sampling strategies such as undersampling frequent words, in \cite{Word2Vec}, are crucial. 
\cite{QuickThoughts} propose contrastive NSP, to predict the next sentence as a positive sample against $n$ random negative sample sentences. Instead of generating the next sentence, they learn to discriminate which sentence encoding follows a given sentence. This allows them to train a better text encoder model with less computation, but sacrifices the ability to generate text. 
\cite{ROBERTA} investigate variations of the contrastive NSP objective used in the BERT model. The method contrasts a consecutive sentence as a positive text sample against multiple non-consecutive sentences from other documents as negative text samples. They find that sampling negatives from the same document during self-supervised BERT pretraining is critical to downstream performance, but that removing the original BERT NSP task improves downstream performance. \cite{CONPONO} find that predicting surrounding sentences in a $k$-sized window around a given central anchor sentence ``improves discourse performance of language models''. 
They sample surrounding sentences: (a) randomly from the corpus to construct easy negatives, and (b) from the same paragraph, but outside the context window as hard (to contrast) negative samples. Contextual negative sampling is theoretically and empirically proven by \cite{ContrastiveLearningLimitations}, who demonstrate that: ``increased negative sampling only helps if negatives are taken from the original texts' context or block of information'', i.e. the same document, paragraph or sentence.
\cite{LM_PRETEXT_TASKS} study how to combine different variants of the NSP \pting tasks with non-contrastive, auxiliary self-supervision signals, while \cite{SentenceStructurePT} explore contrastive sentence structure learning.

\paragraph{Input-output contrastive text representation \pting:}
In \cref{fig:input-output} \cite{rethmeier2020longtail} use output label embeddings as an alternative view $Y$ (labels) of text input embeddings $X$ for contrastive learning of (dis)-similar text-label embedding pairs $(X,Y)$ via binary NCE from \cref{sec:NCE}. Using a separate label and text encoder allows them to efficiently compute many negative label samples, while encoding the text $X$ \emph{only once}, unlike input-input view methods in \cref{fig:input-input}. They pretrain with random input words as pseudo-labels for self-supervised \pting on a very small corpus, which despite the limited \pting data enables unsupervised zero-shot prediction, largely improved few-shot and markedly better rare concept (long-tail) learning.  

\paragraph{Distillation:}
\cite{CT_DISTILL} propose CoDIR, a contrastive language model distillation method to pretrain a smaller student model from an already pretrained larger teacher such as a Masked Transformer Language Model. Compressing a \pted language model is challenging because nuances such as interactions between the original layer representation are easily lost -- without noticing. For distillation, they extract layer representations from both the large teacher and the small student network over the same or two different input texts, to create a student and teacher view of said texts. Using the constrastive InfoNCE loss \cite{Oord}, they then learn to make the student representation similar to teacher representations for the same input texts, and dissimilar if they receive different texts. The score or similarity function in InfoNCE is measured as the cosine distance between mean pooled student and teacher Transformer layer representations. For negative sampling in \pting, they use text inputs from the same topic, e.g.\ a Wikipedia article, to mine hard negative samples -- i.e.\ they sample views from similar contexts as recommended for contrastive methods in \cite{ContrastiveLearningLimitations}.

\paragraph{Text generation as a discriminative EBM:}
\cite{CONTRASTIVE_TEXT_GENERATION} combine an auto-regressive language model, with a contrastive text continuation EBM model for improved text generation. During \pting, they learn to contrast real data text continuations and language model generated text continuations via conditional NCE from \cref{sec:NCE}. For generation, they sample the top-k text completions from the auto-regressive language model and then score the best continuation via the trained EBM, to markedly improve model perplexity. However, the current approach is computationally expensive. 

\paragraph{Cross-modal contrastive representation \pting:}
Representations for zero-shot image classification can be \pted using image caption text for contrastive self-supervised \pting. \cite{ALIGN} automatically mine a large amount of noisy text captions for images in ALIGN, to then noise-filter and use them to construct matching and mismatching pairs of image and augmented text captions for contrastive training. \cite{CLIP} use the same idea in CLIP, but pretrain on a large collection of well annotated image caption datasets. Both methods allow for zero-shot image classification and image-to-text or text-to-image generation, and are inherently zero-shot capable. \cite{CLIP} also run a zero-shot learning efficiency analysis for CLIP and find two things. First, that using a data efficient CNN text encoder increases zero-shot image prediction convergence 3-fold compared to a Transformer text encoder, which they state to be computationally prohibitive. Second, they find that adding contrastive self-supervised text \pting increases zero-shot image classification performance 4-fold. Thus, CLIP \cite{CLIP} shows that contrastive self-supervised CNN text encoder \pting can substantially outperform current Transformer \pting methods, while ALIGN \cite{ALIGN} also automates the image and caption data collection process to increase data scalability.

\subsection{Supervised Contrastive \Pting}
\paragraph{Input-input contrastive supervised text representation \pting}
\cite{GILE} train a two-input-lane Siamese CNN network, which encodes text as the input view $x_i$ in one lane, and labels via a label encoder in a second data view $y_c$, to learn to contrast pairs of $(x_i,y_x)$ as similar (1) or not (0). Rather than encoding labels as multi-hot vectors such as $[0,1,0,0,1]$, they express each label by a textual description of said label. These textual label descriptions can then be encoded by a label encoder subnetwork, which in the simplest case constructs a label embedding by averaging over the word embeddings of the words that describe a label. However, this requires manually describing each label. Using embeddings of supervised labels, they pretrain a contrastive text classification network on known positive and negative labels, and later apply the pretrained network to unseen classes for zero-shot prediction. Their method thus provides supervised, but zero-shot capable \pting. While \cite{rethmeier2020longtail} also support supervised contrastive input-output \pting, they automate label descriptions construction, and conjecture that in real-world scenarios, most labels, e.g. the word `elephant', are already part of the input vocabulary and can thus be pretrained as word embeddings via methods such as Word2Vec \cite{Word2Vec}. They also note that: ``once input words are labels, one can sample input words as pseudo label embeddings for contrastive self-supervised \pting'', as described in section \cref{sec:SSL_CL}. Either method is contrastively pretrained via binary NCE as described in \cref{sec:NCE}. Furthermore, both methods markedly boost few-shot learning and enable zero-shot predictions, while \cite{rethmeier2020longtail} enables unsupervised zero-shot learning via self-supervised contrastive \pting. The added contrastive self-supervision further boosts few-shot and long-tailed learning performance, while also increasing convergence speed over supervised-only contrastive learning in \cite{GILE}.

\paragraph{Contrasting input views on manual text augmentation:}
\cite{NABIKLEIN} use contrastive self-supervised \pting to refine a pretrained BERT language model to drastically increase performance on pronoun disambiguation and the Winograd Schema Commonsense Reasoning task. Their method contrasts over candidate trigger words that affect which word a pronoun refers to. They first mine trigger word candidates from text differences in paraphrased sentences and then maximize the contrastive margin between candidate pair likelihoods. This implicitly pretrains a model for common sense concepts, and is similar to contrastive self-supervision in vision \cite{BIGSIMCLR}, with the difference of the latter generating contrastable data augmentations for a given sample.
While general \pting provides little pronoun disambiguation learning signal, their method demonstrate the design of task-specific contrastive learning to produce strong performance increases in \emph{un- and supervised commonsense reasoning}.

\paragraph{Contrastive text summarization:} \cite{CT_text_summarization} use a Transformer attention mechanism during abstractive sentence summarization learning to optimize two contrasting loss objectives. One loss maximizes the contributions of tokens with the most attention when predicting the summarized sentence. The other loss is connected to a second decoder head, which learns to minimize the contribution of the attention to other, non-summarization relevant, tokens. This method can perhaps best be understood as contrastive, layer attention noise reduction. The main draw back of this method is the current dual network head prediction, which introduces a larger complexity compared to other contrastive methods.

\paragraph{Cross and multi-modal supervised contrastive text \pting for representation learning:}
Recent work from computer vision and time series prediction train with contrastive supervised losses to enable zero-shot learning or improve data-to-text generation. \cite{Vision_Text_CP} fuse image an text description information into the same representation space for generalized zero-shot learning -- i.e.\ where at test time some classes are unseen, zero-shot, while other classes were seen during training. To do so, they first pretrain a supervised text-image encoder network to contrast $(image, text, label)$ triplets of human annotated image classes. At test time, this contrastive network decides which text description best matches a given image. This works for seen and unseen classes, because classes are represented as text descriptions. \cite{CLIP} pretrains on manually annotated textual image descriptions to enable better generalization to unseen image classes.
\cite{DATA_TO_TEXT_GENERATION} turn stock price value time series into textual stock change descriptions where the contrastive objectives markedly increase the fluency and non-receptiveness of generated texts, especially when trained with little data.

\paragraph{Datasets construction for contrastive \pting:}
\cite{MULTI_LING_CONTRASTIVE_WSD_EVAL} automatically create a corpus of contrastive sentences for word sense disambiguation in machine translation by first identifying sense ambiguous source sentence words, and then creating replacement word candidates to mine sentences for contrastive evaluation. 

\section{Challenges and Potential Directions}
\paragraph{Challenge: need for many negatives.} Current methods require the sampling of many negative instances for contrastive learning to work well. There is work on the benefits and harms of sampling hard to contrast negatives \cite{hard_negatives}, or relevant negatives \cite{ContrastiveLearningLimitations}, which can boost sampling efficiency. However, as seen in \cite{W2V,rethmeier2020longtail} depending on the task, sampling diverse negatives can play an important role. To date, the importance of easy to contrast negative samples is underexplored, but insights from a metric learning survey by \cite{MetricLearningProblems}, suggest that hard, medium and easy samples may all be necessary, especially for generalization in open class set tasks such as \pting.  

\paragraph{Challenge and directions: text augmentation quality and efficiency:}
Self-supervised text augmentation research in NLP (\cref{sec:SSL_CL}) is gaining momentum and \cite{ContrastiveDataAugmentationPretraining,textDataAugmentations} and many others analyze using mixes of recent text data augmentations. However, these input-input contrastive methods often use computationally expensive or non-robust mechanisms like: back translation, initializing a new prediction head per downstream task, or reliance on already otherwise \pted models like RoBERTa. Works on input-output contrastive learning like \cite{GILE,rethmeier2020longtail} nullify these requirements and demonstrate very data efficient \pting, which is currently an under-researched, but very desirable property of contrastive learning. \cite{CLT_INVERTS_DATA_GENERATION} further solidify these insights and show that contrastive methods effectively recover data properties even from small data sets. While many self-supervised contrastive \pting methods rely on already \pted Transformers, works \cite{rethmeier2020longtail,EBM_LM_PRETRAINING,CLEAR,COCOLM} make important contributions by removing this restriction. \cite{CLEAR,CONPONO} propose robustly scalable input augmentation, while \cite{BYOL} propose BYOL, which does not require negative sampling, and potentially lends itself improving to future contrastive NLP methods.

\paragraph{Challenge: under-researched applications:} \cite{CONTRASTIVE_TEXT_GENERATION} enhance a text generation language model with contrastive importance resampling of language model generated text continuations. \cite{CT_text_summarization} propose contrastive abstractive sentence summarization, which using Momentum Contrast can potentially improve on. 

\paragraph{Direction: cross-modal generation:}
An underresearch direction for contrastive NLP are data-to-text tasks that turn non-text inputs into a textual description. For example \cite{DATA_TO_TEXT_GENERATION} contrastively learn to generate stock change text descriptions from stock price time series using limited data, while works like \cite{CLIP,ALIGN} show that contrastive text supervision and self-supervision can multiply the zero-shot learning efficiency in cross-modal representation learning. 

\paragraph{Direction: contrastive (language) model fusion:}
While \cite{CT_DISTILL} compress a large language model, which future work can adapt to fuse multiple language model or mutually transfer knowledge between models.

\paragraph{Direction: commonsense contrastive learning:}
The contrastive word sense disambiguation (WSD) dataset construction method by \cite{MULTI_LING_CONTRASTIVE_WSD_EVAL} is potentially adaptable to automatically mine inputs for the contrastive pronoun learning method by \cite{NABIKLEIN}. 

\section{Conclusion}
In this primer on contrastive \pting, we surveyed contrastive learning concepts and their relations to other fields. 
We also structured contrastive \pting as self- vs. supervised learning, highlighted existing challenges and provided pointers to future research directions.     

\bibliographystyle{named}
\bibliography{bib}

\end{document}

%% file: math_commands.tex
\usepackage{amsmath,amsfonts,bm}

\def\eqref#1{equation~\ref{#1}}

\def\1{\bm{1}}

\DeclareMathAlphabet{\mathsfit}{\encodingdefault}{\sfdefault}{m}{sl}
\SetMathAlphabet{\mathsfit}{bold}{\encodingdefault}{\sfdefault}{bx}{n}

